\documentclass[sigconf,screen=true,anonymous=false,bookmarks=false,nonacm]{acmart}
\iftrue
\usepackage{titlesec}
\titlespacing\section{2pt}{5pt plus 1pt minus 1pt}{0pt plus 1pt minus 1pt}
\titlespacing\subsection{2pt}{5pt plus 1pt minus 1pt}{0pt plus 1pt minus 1pt}
\titlespacing\subsubsection{2pt}{5pt plus 1pt minus 1pt}{2pt plus 1pt minus 1pt}
\usepackage{enumitem}
\setlist{leftmargin=5.08mm}
\fi

\iftrue
\fi

\usepackage{lipsum}
\usepackage{graphicx}
\usepackage{amsmath}
\usepackage{footnote}
\usepackage{mathtools}
\usepackage{mathrsfs}     
\usepackage{comment}
\usepackage[subrefformat=parens,labelformat=parens]{subfig}
\usepackage{bm,bbm}
\usepackage{multirow}
\usepackage{threeparttable,booktabs}
\usepackage{blkarray}
\usepackage{tikz}
\usetikzlibrary{positioning,calc,fit,decorations.pathmorphing,shapes.geometric,shapes.gates.logic.US,calc}
\usepackage{tikz-3dplot}
\usepackage{balance}
\usepackage{courier}                                       
\usepackage[mathcal]{eucal}
\usepackage[]{algpseudocode}                               
\algrenewcommand\textproc{\texttt}
\makeatletter\let\float@addtolists\relax\makeatother
\usepackage{algorithm}
\renewcommand{\algorithmicrequire}{\textbf{Input:} }       
\renewcommand{\algorithmicensure} {\textbf{Output:}}       
\usepackage{cleveref}                                      
\usepackage{filecontents}                                  
\usepackage{pgfplots}
\usepackage{pgfplotstable}
\usepackage{pgf-pie}
\usepgfplotslibrary{groupplots}
\pgfplotsset{compat=newest}
\usepackage[figuresright]{rotating}
\usepackage{xcolor,colortbl}                               

\newcommand{\minisection}[1]{\vspace{.06in}\noindent{\textbf{#1}}.}

\theoremstyle{plain}

\theoremstyle{definition}

\algrenewcommand\textproc{\texttt}

\usepackage[skip=1pt]{caption}            
\definecolor{CUHKorange}{RGB}{244,106,18} 
\definecolor{CUHKblue}{RGB}{0,111,190}    
\definecolor{CUHKgreen}{RGB}{0,127,128}   
\definecolor{CUHKred}{RGB}{228,46,36}     
\definecolor{CUHKyellow}{RGB}{198,148,34} 
\definecolor{CUHKdark}{RGB}{114,44,114}   
\definecolor{CUHKmiddle}{RGB}{144,44,144} 

\usepackage{tcolorbox}
\tcbuselibrary{skins,breakable}
    {\endtcolorbox}

\graphicspath{{./figs/}}
\setcopyright{none}

\usepackage{algpseudocodex}

\usepackage{makecell}

\begin{document}
\date{}

\title{Differentiable Edge-based OPC}

\author{Guojin Chen}
\affiliation{
    \institution{CUHK \& UT Austin \& NVIDIA}
}
\authornote{Work accomplished during internship at NVIDIA.}
\author{Haoyu Yang}
\affiliation{
    \institution{NVIDIA Corporation}
}

\author{Haoxing Ren}
\affiliation{
    \institution{NVIDIA Corporation}
}

\author{Bei Yu}
\affiliation{
    \institution{Chinese University of Hong Kong}
}

\author{David Z. Pan}
\affiliation{
    \institution{University of Texas at Austin}
}

\begin{abstract}
Optical proximity correction (OPC) is crucial for pushing the boundaries of semiconductor manufacturing and enabling the continued scaling of integrated circuits. While pixel-based OPC, termed as inverse lithography technology (ILT), has gained research interest due to its flexibility and precision. Its complexity and intricate features can lead to challenges in mask writing, increased defects, and higher costs, hence hindering widespread industrial adoption. In this paper, we propose DiffOPC, a differentiable OPC framework that enjoys the virtue of both edge-based OPC and ILT. By employing a mask rule-aware gradient-based optimization approach, DiffOPC efficiently guides mask edge segment movement during mask optimization, minimizing wafer error by propagating true gradients from the cost function back to the mask edges.
Our approach achieves lower edge placement error while reducing manufacturing cost by half compared to state-of-the-art OPC techniques, bridging the gap between the high accuracy of pixel-based OPC and the practicality required for industrial adoption, thus offering a promising solution for advanced semiconductor manufacturing.
\end{abstract}
\maketitle
\pagestyle{empty}

\section{Introduction}
\label{sec:intro}

Optical proximity correction (OPC) is a critical technique in computational lithography that compensates for the optical proximity effect (OPE) caused by interference and diffraction in the lithographic imaging process. As integrated circuit technology nodes advance to 90 nm and below, simple resolution enhancement techniques (RET) can no longer meet the requirements for high-resolution and high-fidelity lithographic imaging. To address this challenge, OPC has evolved from rule-based OPC (RBOPC) to model-based OPC (MBOPC).

RBOPC relies on a pre-established mask correction rule table, which is derived from engineering experience or fitted experimental and simulation data~\cite{OPC-SPIE1994-Otto}. Although RBOPC is computationally fast and produces relatively simple optimized mask patterns, it can only compensate for local OPE and cannot find a globally optimal solution for the mask optimization problem.

\begin{figure}[!tbp]
  \centering
  \includegraphics[width=0.82\linewidth]{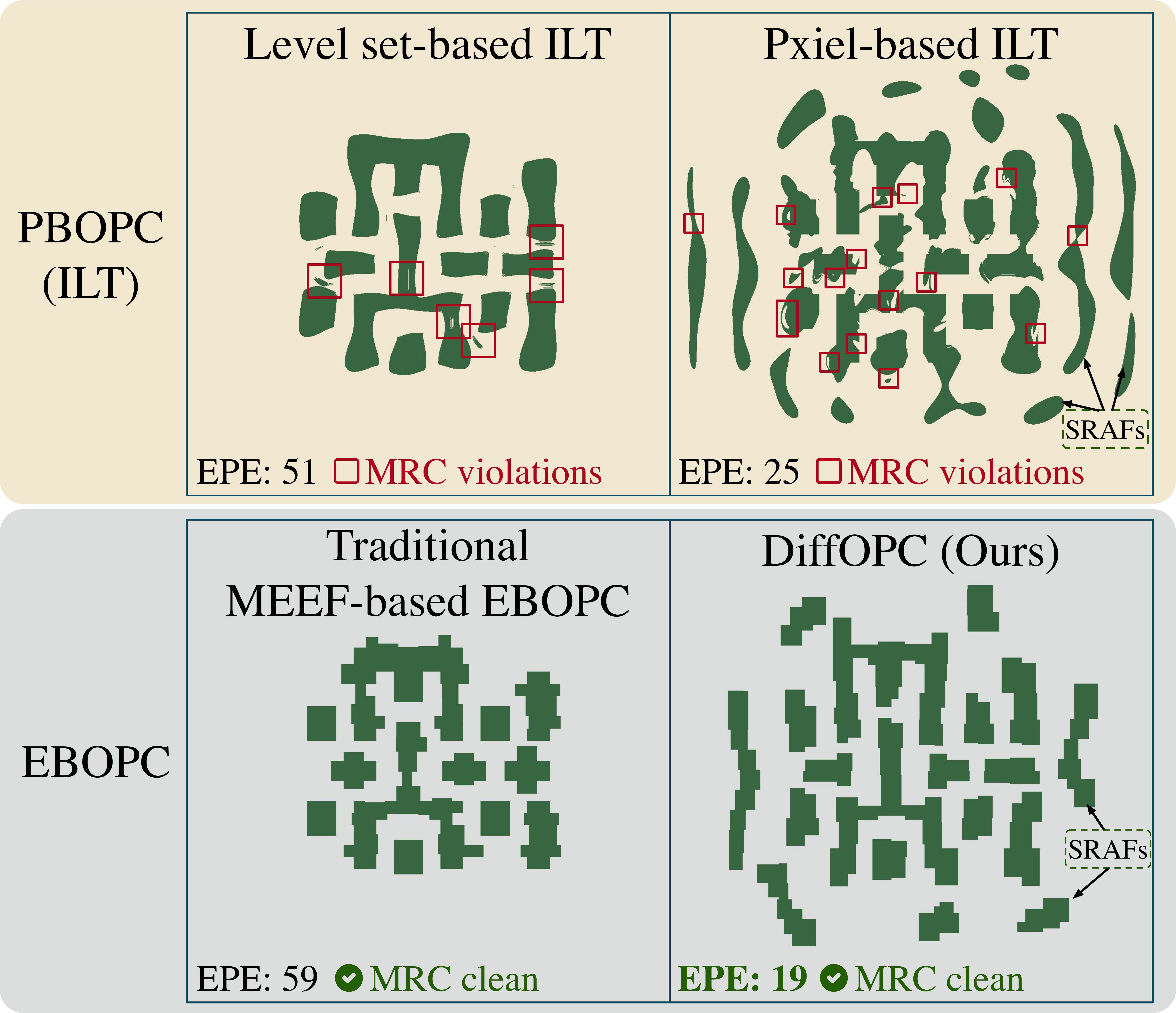}
  \caption{Model-based OPC includes pixel-based OPC (ILT) and edge-based OPC (EBOPC). While ILT masks face manufacturability issues requiring significant post-processing, EBOPC masks are manufacturable but have performance limitations. DiffOPC combines the advantages of both approaches, enhancing manufacturability and performance.}
  \label{fig:overview}
  \vspace{-.8em}
\end{figure}

MBOPC, on the other hand, is based on the physical model of lithographic imaging and employs numerical optimization algorithms to modify the mask pattern. As depicted in \Cref{fig:overview}, MBOPC can be further classified into edge-based OPC (EBOPC) and pixel-based OPC (PBOPC). EBOPC divides the edge contour of the mask pattern into several segments and iteratively optimizes the position of each segment along its normal direction to compensate for lithographic imaging errors~\cite{MEEF-lei2014model}.

However, current EBOPC methods, such as the Mask Error Enhancement Factor (MEEF) matrix algorithm~\cite{MEEF-lei2014model}, have limitations in computational efficiency and accuracy. The algorithm is computationally intensive, scaling poorly with the size and complexity of IC layouts. Its foundational linearity assumptions often fail to account for the nonlinearities prevalent in advanced lithography, leading to subpar performance in complex cases where edge interactions are significant and not adequately captured. The MEEF matrix, further burdened by potential ill-conditioning and a static representation throughout optimization, may not adapt to dynamic process variations, thus trading off accuracy for computational manageability.

PBOPC, also known as inverse lithography technology (ILT), pushes the boundaries of mask optimization by rasterizing the mask layout into a pixel array and optimizing the transmission of each mask pixel by gradient descent~\cite{OPC-DAC2014-Gao}.
This approach allows for free-form curved edge contours and the addition of sub-resolution assist features (SRAF)~\cite{SRAF-brist2003model,OPC-TCAD2017-Xu,OPC-DAC2019-Alawieh,OPC-TCAD2020-Geng} to improve imaging performance.
ILT algorithms can be categorized into two classes based on their mask representation:
end-to-end pixel-based methods for prediction~\cite{OPC-DAC2018-Yang,OPC-ICCAD2020-DAMO,OPC-ICCAD2020-NeuralILT,ICCAD22_AdaOPC,opc-TCAD2023-L2O-ILT,zhang2024fracturing,OPC-multiilt2023sun} or acceleration,
and implicit function-based methods using level sets to enhance acceleration and manufacturability~\cite{OPC-DATE2021-Yu,OPC-ICCAD2021-DevelSet,OPC-JVSTB2013-Lv}.
Among the SOTA ILT methods, MultiILT~\cite{OPC-multiilt2023sun} adopts a multi-level resolution strategy for better OPC performance and manufacturability.

Despite the advancements in ILT algorithms, they still face several challenges that hinder their widespread adoption in the semiconductor industry.
As illustrated in \Cref{fig:overview}, the pixelated mask patterns generated by ILT are often complex and difficult to manufacture, requiring costly rectangular decomposition into manufacturable Manhattan polygons.
Further, the application of decomposition and mask rule check (MRC) methods to regularize the mask patterns may lead to a decline in OPC performance and introduce new hotspots, negating the performance advantages of ILT.
Moreover, ILT algorithms tend to over-optimize shape corners because the simulated line-ends will never match the Manhattan rectangles at the line-end.
Nevertheless, these challenges have been largely overlooked, preventing ILT's large-scale adoption in the industry, which tends to favor EBOPC due to lower manufacturing costs.

To bridge the gap between the manufacturability of EBOPC and the performance of ILT, we propose DiffOPC, a differentiable edge-based OPC method that leverages gradient information to optimize edge placement error (EPE) while considering process variation. By relaxing discrete edge movements and embedding mask rule constraints into the gradient computation, DiffOPC combines EBOPC's high manufacturability with ILT's performance. Additionally, it ensures MRC-clean results, allowing the optimized mask patterns to be directly used for mask fabrication without additional post-processing.

DiffOPC introduces efficient solutions to enhance the edge-based OPC process.
In the forward algorithm, a flexible segmentation approach and CUDA-accelerated ray casting expedite differentiable layout rasterization, while a novel SRAF seed generation algorithm optimizes SRAF placement.
In the backward algorithm, DiffOPC computes lithography gradients for edge movements using a chain-rule approach and incorporates mask rule constraints to ensure manufacturability. By combining these improvements, DiffOPC achieves superior OPC performance with high manufacturability.
In summary, our main contributions are as follows:
\begin{itemize}
    \item We propose DiffOPC, a differentiable edge-based OPC framework that integrates EPE loss and leverages MRC-aware gradients for mask optimization.
    \item A flexible segmentation approach and a CUDA-accelerated ray casting algorithm are introduced to expedite layout rasterization. 
    \item DiffOPC efficiently computes edge segment gradients using a chain-rule approach to ensure manufacturability.
    \item A novel SRAF seed generation algorithm leveraging gradients for optimal SRAF placement and further optimization.
    \item DiffOPC bridges the gap between EBOPC's manufacturability and ILT's performance, offering a promising solution for high-quality and efficient OPC corrections.
    The experimental results show that DiffOPC reduces EBOPC's EPE by half, and even achieves lower EPE than ILT while maintaining manufacturing costs that are half of ILT's.
\end{itemize}

\section{Preliminaries}
\label{sec:prelim}

\subsection{Forward Lithography Model}
We employ the sum of coherent systems (SOCS) decomposition of a 193$nm$ wavelength system as the optical model for lithography modeling, following the same approach as ~\cite{OPC-ICCAD2013-Banerjee}.
The aerial image intensity $\boldsymbol{I}$ is represented by the convolution of the mask $\boldsymbol{M}$ and a set of optical kernels $\boldsymbol{H}$.
The $N_k^{th}$ order approximation to the partially coherent system is obtained using \cref{eq:approx_intensity}:
\begin{equation}
\boldsymbol{I}(x, y) \approx \sum_{i=1}^{N_{k}} \sigma_{i}\left|\boldsymbol{M}(x, y) \otimes h_{i}(x, y)\right|^{2},
\label{eq:approx_intensity}
\end{equation}
where $\otimes$ denotes the convolution operation, $h_i$ is the $i^{th}$ kernel of $\boldsymbol{H}$, $\sigma_i$ is the corresponding weight of the coherent system, and $(x,y)$ is the index notation of the matrix. $\boldsymbol{M}(x, y)$ represents the pixel value at the point $(x, y)$ of the mask image $\boldsymbol{M}$.
A  constant threshold resist model (CTR) is applied to convert the aerial image intensity $\boldsymbol{I}$ to the printed resist image $\boldsymbol{Z}$.
\begin{equation}
\begin{aligned}
    \boldsymbol{Z}(x, y) = \left\{
        \begin{array}{ll}
            1, & \mbox{if $\boldsymbol{I}(x, y) > I_{th}$, } \\
            0, & \mbox{otherwise, }
        \end{array}
        \right.
    \label{eqn:resist1}
\end{aligned}
\end{equation}
where $I_{th}$ is the intensity threshold.

\subsection{Evaluation Metrics}
In this paper, we use squared $L_2$ error, process variation band (PVB) and edge placement error (EPE) as three typical metrics to evaluate OPC performance.
Moreover, the mask fracturing shot count (\#shot) proposed in~\cite{OPC-ICCAD2020-NeuralILT} is also applied in this work to evaluate mask complexity and manufacturability.
\begin{description}
    \item[Squared $L_2$ error] $L_2$ measures the difference between the nominal resist image $\boldsymbol{Z}_{nom}$ and the target image $\boldsymbol{T}$, defined as:
        \begin{equation}
            \mbox{$L_2$}(\boldsymbol{Z}_{nom}, \boldsymbol{T}) = \Vert \boldsymbol{Z}_{nom} - \boldsymbol{T} \Vert_2^2.
            \label{eqn:l2loss}
        \end{equation}
    \item[PVB] evaluates the robustness of the mask against different process conditions.
        A smaller PVB indicates a more robust mask.
        $\mbox{PVB}(\boldsymbol{Z}_{max}, \boldsymbol{Z}_{min}) = \Vert \boldsymbol{Z}_{max} - \boldsymbol{Z}_{min} \Vert_2^2.   \label{eqn:pvb}$
    \item[Edge placement error] The Edge Placement Error (EPE)~\cite{OPC-ICCAD2013-Banerjee} quantifies the geometric distortion of the resist image. 
    \item[Shot count]  \#Shot~\cite{OPC-ICCAD2020-NeuralILT} is the number of decomposed rectangles that replicate the original mask exactly.
\end{description}

\subsection{Problem Formulation}

Given a target design $\boldsymbol{T}$,
we aim to find a set of boundary segments $\boldsymbol{S} = \{\boldsymbol{s}_1, \boldsymbol{s}_2, \ldots, \boldsymbol{s}_i\}$, and a binary mask $\boldsymbol{M} \in \{0, 1\}^{m \times n}$ formed by the matrix inside the boundary composed of these segments $\boldsymbol{S}$, where $m$ and $n$ are the dimensions of $\boldsymbol{T}$. The objective is to determine the corresponding printed image $\boldsymbol{Z}$ that minimizes the weighted sum of EPE, $L_2$, PVB, and \#shots.

\section{DiffOPC Algorithm}

\label{sec:algo}

\begin{figure*}
  \centering
  \includegraphics[width=0.78\textwidth]{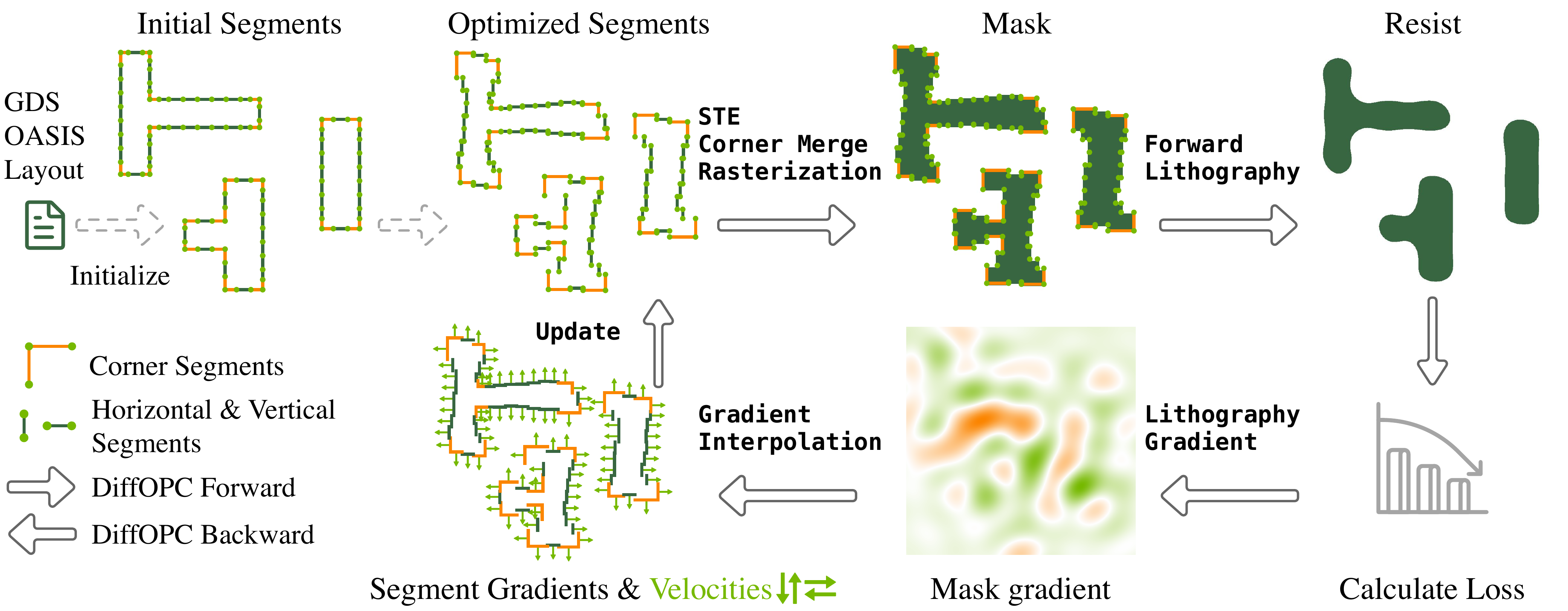}
  \caption{DiffOPC: differentiable edge-based OPC framework.}
  \label{fig:diffopc_flow}
\end{figure*}

To enable the application of differentiable EBOPC to arbitrary layout patterns while utilizing minimal additional information, such as the EPE measure points, several challenges need to be addressed:
1) Ensuring a more flexible movement of segments in Manhattan geometries, particularly at pattern corners.
2) Mapping discrete edge movements to a continuous space for efficient updates.
3) Maintaining compatibility with the chain rule for differentiation during the rasterization process, which converts edge parameters to pixel binary masks.
In this section, we introduce the movement and update mechanisms for edge segments, describe a CUDA-accelerated ray casting algorithm for rasterization, demonstrate how lithography gradients can be utilized to update the movement of edge segments, and introduce an algorithm for SRAF placement.

\subsection{Edge Segmentation and Movement}
We present \Cref{alg:edge_init} for segmenting target polygon edges into smaller segments of a pre-defined length.
The algorithm returns a minimal set of segments, denoted as $\boldsymbol{S} \in \mathbb{R}^{N_s \times 2 \times 2}$, where $\mathbb{R}$ represents the real number domain and $N_s$ is the number of segments.
Each segment $\boldsymbol{s}_i \in \boldsymbol{S}$ is represented by its starting and ending coordinates in vector form: $[[x_1, y_1], [x_2, y_2]]$.
These segments $\boldsymbol{S}$ serve as the optimization parameters for DiffOPC, providing increased flexibility in handling corner edges compared to traditional EBOPC methods which only optimize the edge movement distance.
As illustrated in \Cref{fig:diffopc_flow}, each segment $\boldsymbol{s}_i$ is associated with a direction vector $\boldsymbol{d}_i \in \mathcal{D}$, which enables better reconstruction of segments back into polygons and determines the direction of movement. Furthermore, the algorithm ensures compliance with the MRC by merging excessively short segments when necessary.
\begin{algorithm}
  \caption{Edge Parameter Initialization.}
  \small
  \begin{algorithmic}[1]
  \Require {$polygons$: mask polygon coordinates; $seg\_length$: segment length}.
  \Ensure $all\_segments$: List of polygons with segmented lines \& directions;
  \For{$poly$ in $polygons$}
  \For{$edge$ in $poly$}
  \State $midpoint \gets$ Calculate the midpoint of the edge;
  \State $length \gets$ Calculate the length of the edge;
  \State $direction \gets$ Get edge direction vector (horizontal or vertical);
  \If{$length \leq 2 * seg\_length$}
  \State Create two segments from $midpoint$;
  \Else
  \State $steps \gets$ Calculate the number of steps based on edge $length$ and $seg\_length$;
  \For{$i \gets -steps$ to $steps$}
  \State Calculate the start and end points of the segment based on $midpoint$ and step size;
  \If{segment length $>$ $seg\_length$}
  \State Split the segment into two segments at the $midpoint$;
  \Else
  \State Create a single segment;
  \EndIf
  \EndFor
  \EndIf
  \State Mark the start or the end of edge segments as corner segments;
  \State Add the segments and directions to polygon segments list;
  \EndFor
  \State Add the polygon segments and directions to $all\_segments$;
  \EndFor
  \State \Return $all\_segments$;
  \end{algorithmic}
  \label{alg:edge_init}
\end{algorithm}

In DiffOPC, after determining the segments $\boldsymbol{S}$ and their corresponding directions $\mathcal{D}$, it is crucial to establish the velocity vector $\boldsymbol{v}_i$ for each segment $\boldsymbol{s}_i$. The velocity vector connects the movement of edge segments with the gradients obtained from lithography simulations.
The concept of velocity vectors is inspired by level set-based ILT (LSILT).
In LSILT, the velocity component is the projection of the gradient of the implicit level set function $\phi$ onto the mask plane, denoted as $\nabla \phi$, which can be a vector in any direction.
However, in DiffOPC, the movement direction $\boldsymbol{v}_i$ of an edge segment $\boldsymbol{s}_i$ is restricted to be perpendicular to its direction vector $\boldsymbol{d}_i$, (either horizontal or vertical), satisfying the condition $\boldsymbol{v}_i \cdot \boldsymbol{d}_i = 0$.
Additionally, we set the default orientation of all velocity vectors $\boldsymbol{v}_i$ to point outward from the polygon, as illustrated in \Cref{fig:diffopc_flow}.

\subsection{Differentiable Edge-Based OPC}
The preprocessed data consists of segments $\boldsymbol{S}$ and corresponding velocity vectors $\boldsymbol{V}$. $\boldsymbol{S}$ is stored as learnable parameters in tensor $\boldsymbol{S} \in \mathbb{R}^{N_s \times 2 \times 2}$, while $\boldsymbol{V}$ is a fixed tensor $\boldsymbol{V} \in \mathbb{R}^{N_s \times 2}$ used in computations, where $N_s$ is the number of segments.
In DiffOPC, the forward pass from edge parameters $\boldsymbol{S}$ to the resist image $\boldsymbol{Z}$ involves five differentiable steps:
1) Edge parameter rounding.
2) Merging corner edges.
3) Edge-to-mask rasterization.
4) Forward lithography simulation.
5) Loss calculation.
Each step's forward and backward computations will be discussed in detail in this chapter.

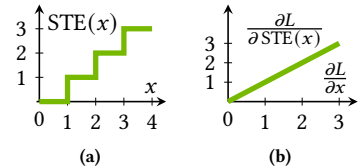
\begin{figure}[h]
  \vspace{-.5em}
  \centering
  \subfloat[]{ \definecolor{ICCAD13Color}{HTML}{76b900}
\begin{tikzpicture}[
        font=\normalsize,
        declare function={
          f1(\x)=(
          \x*exp(2*\x/25)/(1+exp(8)+exp(2*\x/25))-\x*exp(-2*\x/25)/(1+exp(-10)+exp(-2*\x/25))+
          100*exp(8)/(1+exp(8)+exp(2*\x/25))-100*exp(-10)/(1+exp(-10)+exp(-2*\x/25))
          );
          f2(\x)=(
          \x*exp(\x/20)/(1+exp(5)+exp(\x/20))-\x*exp(-\x/20)/(1+exp(-5)+exp(-\x/20))+
          100*exp(5)/(1+exp(5)+exp(\x/20))-100*exp(-5)/(1+exp(-5)+exp(-\x/20)) 
          );
          f3(\x)=round(\x);
        },
        line/.style={line width=2pt,blue!50}
      ]
      \definecolor{Blue-d}{HTML}{3a76af}
      \begin{axis}[
          scale only axis,clip=false,
          at={(0,0)},
          width=0.2\linewidth,
          height=0.15\linewidth,
          axis line style={line width=.6pt},
          tick align=outside,
          every major tick/.append style={ %
            major tick length=3pt, black},
          legend cell align=left,
          xlabel style={font=\normalsize},
          x tick label style={
            /pgf/number format/assume math mode, font=\normalsize},
          xmin=-.1,xmax=4.5,ymax=4,
          axis lines = middle,
          xlabel=$x$,ylabel=$\operatorname{STE}({x})$,
          xtick={0,1,2,3, 4},
          ytick={0,1,2,3},
          ytick distance=2,
          y tick label style={
            /pgf/number format/assume math mode, font=\normalsize},
          legend style={
            draw=none, at={(0.5,1)},
            anchor=north, fill=none,
            nodes={scale=0.9, transform shape}
          },
          every axis plot/.append style={tension=0,line width=1.2pt}
        ]
        \addplot[domain=0:3,smooth,draw=none] {f3(x)};
        \node[label={[label distance=-1mm]-90:0}] at (axis cs:0,0) {};
        \draw[line,color=ICCAD13Color] (0,0)--(1,0)--(1,1)--(2,1)--(2,2)--(3,2)--(3,3)--(4,3);
      \end{axis}
    \end{tikzpicture} \label{fig:ste-forward} }
  \subfloat[]{ \definecolor{ICCAD13Color}{HTML}{76b900}
\begin{tikzpicture}[
        font=\normalsize,
        declare function={
          f1(\x)=(
          \x*exp(2*\x/25)/(1+exp(8)+exp(2*\x/25))-\x*exp(-2*\x/25)/(1+exp(-10)+exp(-2*\x/25))+
          100*exp(8)/(1+exp(8)+exp(2*\x/25))-100*exp(-10)/(1+exp(-10)+exp(-2*\x/25)) 
          );
          f2(\x)=(
          \x*exp(\x/20)/(1+exp(5)+exp(\x/20))-\x*exp(-\x/20)/(1+exp(-5)+exp(-\x/20))+
          100*exp(5)/(1+exp(5)+exp(\x/20))-100*exp(-5)/(1+exp(-5)+exp(-\x/20)) 
          );
          f3(\x)=round(\x);
        },
        line/.style={line width=2pt,blue!50}
      ]
      \definecolor{Blue-d}{HTML}{3a76af}
      \begin{axis}[
          scale only axis,clip=false,
          at={(0,0)},
          width=0.2\linewidth,
          height=0.15\linewidth,
          axis line style={line width=.6pt},
          tick align=outside,
          every major tick/.append style={ %
            major tick length=3pt, black},
          legend cell align=left,
          xlabel style={yshift=1cm},
          x tick label style={
            /pgf/number format/assume math mode, font=\normalsize},
          xmin=0,xmax=3.5,ymax=5,
          axis lines = middle,
          ylabel style={xshift=.1cm},
          xlabel=$\frac{\partial L}{\partial x}$,ylabel=$\frac{\partial L}{\partial \operatorname{STE}({x})}$,
          xtick={0,1,2,3,4},
          ytick={0,1,2,3},
          ytick distance=2,
          y tick label style={
            /pgf/number format/assume math mode, font=\normalsize},
          legend style={
            draw=none, at={(0.5,1)},
            anchor=north, fill=none,
            nodes={scale=0.9, transform shape}
          },
          every axis plot/.append style={tension=0,line width=1.2pt}
        ]
        \addplot[domain=0:3,smooth,draw=none] {f3(x)};
        \node[label={[label distance=-1mm]-90:0}] at (axis cs:0,0) {};
        \draw[line,color=ICCAD13Color] (0,0)--(3,3);
      \end{axis}
    \end{tikzpicture} \label{fig:ste-backward} }
  \caption{(a) STE forward; (b) STE backward.}
  \label{fig:ste-all}
\end{figure}

\minisection{Differentiable edge parameter rounding}
Since the edge parameters $\boldsymbol{S}$ are real-valued, while the edge coordinate system is integer-valued, the rounding operation is non-differentiable. To address this issue and enable a differentiable process, we employ the straight-through estimator (STE) for rounding $\boldsymbol{S}$.
\begin{equation}
  \bar{x}_i = \operatorname{STE}(x_i),~\bar{y}_i = \operatorname{STE}(y_i),~\bar{\boldsymbol{s}_i} = \operatorname{STE}(\boldsymbol{s}_i),~\bar{\boldsymbol{S}} = \operatorname{STE}(\boldsymbol{S}).
\end{equation}
STE is defined as:
\begin{flalign}
  \begin{aligned}
    \bar{x} = \operatorname{STE}(x) &= \text{Round}(x),  &\Comment{STE forward.} \\
    \frac{\partial L}{\partial \operatorname{STE}({x})} &= \frac{\partial L}{\partial x}. &\Comment{STE backward.}
  \end{aligned}
  \label{eq:ste}
\end{flalign}
The \textbf{forward} pass illustrated in \Cref{fig:ste-forward} applies the rounding function to $\boldsymbol{S}$, while the \textbf{backward} pass directly propagates the gradients from $\bar{\boldsymbol{S}}$ to $\boldsymbol{S}$, as shown in \Cref{fig:ste-backward}.

\minisection{Corner edge merging}
During the optimization, as edges move, the endpoints of different segments separate.
For non-corner segments, the segment length remains unchanged since they only move along the normal direction. The newly formed edges between adjacent segments can be obtained from the endpoints of the neighboring edges without additional processing. However, for segments adjacent to corners, the movement directions differ, requiring extra handling. After the movement, the new intersection point may lie outside the two segments. Therefore, it is necessary to additionally connect the segments adjacent to the corners.
The adjusted algorithm is presented in \Cref{alg:corner_merge}.
After the \textbf{forward} pass of the corner merging operation, the modified edge parameters $\hat{\boldsymbol{S}}$ ensure that all adjacent segments at the corners are re-connected.
The \textbf{backward} pass is straightforward, as the gradients are directly propagated to the rounded edge parameters $\bar{\boldsymbol{S}}$.

\begin{algorithm}
  \caption{Find Intersection and Adjust Corner Segments}
  \small
  \begin{algorithmic}[1]
  \Function{FindIntersectionAndAdjust}{$\bar{s}_1, \bar{s}_2$}
      \State $s_v \gets$ vertical segment; $s_h \gets$ horizontal segment;
      \State $p^{\dagger} \gets$ the intersection point ($x$ of $s_v$, $y$ of $s_h$);
      \If{$\bar{s}_1$ is vertical}
          \State Adjust the end point of the vertical line $s_v$ to $p^{\dagger}$;
          \State Adjust the start point of the horizontal line $s_h$ to $p^{\dagger}$;
      \Else
          \State Adjust the end point of the horizontal line $s_h$ to $p^{\dagger}$;
          \State Adjust the start point of the vertical line $s_v$ to $p^{\dagger}$;
      \EndIf
      \State \Return the adjusted segments $(\hat{s}_1, \hat{s}_2)$;
  \EndFunction
  \end{algorithmic}
  \label{alg:corner_merge}
\end{algorithm}

\minisection{Differentiable rasterization using CUDA-accelerated ray casting}
The core challenge in DiffOPC is the edge-to-pixel rasterization, as the lithography model in \cref{eq:approx_intensity} only accepts pixel-based mask input $\boldsymbol{M}$. This rasterization process must be differentiable to allow the gradient flow to reach the edge parameters from the mask, and it should be as fast as possible since it is performed in every optimization epoch. Traditional EBOPC methods involve moving segments and then filling or subtracting the corresponding binary matrix at the new positions. However, this approach is time-consuming due to the need to sequentially access each segment and convert the segment's displacement into mask indices, repeatedly reading and modifying the corresponding locations. To address these issues, a method that effectively generates a binary mask from rounded edge parameters using CUDA-accelerated ray casting is proposed in \Cref{alg:rasterization}.

\begin{algorithm}
  \caption{Parallelized Ray Casting for Edge to Mask Rasterization}
  \begin{algorithmic}[1]
  \Require Merged edge parameters: $\hat{\boldsymbol{S}}$, width $W$, height $H$;
  \Ensure Binary mask: $mask$;
  \Function{\textnormal{\texttt{Rasterize}}}{$\hat{\boldsymbol{S}}, W, H$}
  \State $mask \gets \text{zeros}((W, H))$;~$bbox \gets \text{bounding\_box}(\hat{\boldsymbol{S}})$;
  \State $points \gets \text{grid\_within}(bbox)$;~$count \gets \text{zeros}((W, H))$;
  \State \Comment{Create grid points and initialize count}
  \State $\hat{\boldsymbol{S}}_h \gets$ extract horizontal segments from $\hat{\boldsymbol{S}}$; \label{line:extract_h}
  \For{$\boldsymbol{s}_i$ in $\hat{\boldsymbol{S}}_h$ parallelly} \Comment{Parallel computation}
      \For{$p$ in $points$ parallelly} 
          \State $cross \gets \texttt{check\_cross}(p, \boldsymbol{s}_i)$;
          \State $count[p] \gets count[p] + cross$ \Comment{Accumulate checks}
      \EndFor
      \State \texttt{\_\_syncthreads()}; \Comment{Synchronize threads}  
  \EndFor
  \State $mask \gets \text{mod}(count, 2) == 1$; \Comment{Apply even-odd rule} 
  \State \Return $mask$;
  \EndFunction
  \Function{\textnormal{\texttt{check\_cross}}}{$p$, $\boldsymbol{S}$}
  \State $v1 \gets p - s.start$;~$v2 \gets p - s.end$; \Comment{Vectors from $p$ to $\boldsymbol{S}$}
  \State $cross \gets v1.x \times v2.y - v1.y \times v2.x$; \Comment{Cross product }
  \State $cond1 \gets (v1.x < 0) \text{ and } (v2.x \geq 0) \text{ and } (cross < 0)$;
  \State $cond2 \gets (v1.x \geq 0) \text{ and } (v2.x < 0) \text{ and } (cross > 0)$;
  \State \Return $cond1$ or $cond2$; \Comment{True if ray crosses the segment}
  \EndFunction
  \end{algorithmic}
  \label{alg:rasterization}
\end{algorithm}
\Cref{alg:rasterization} presents an efficient, fully parallelized method for generating a binary mask from edge parameters using ray casting. The main function, \texttt{Rasterize}, initializes an empty mask and a count matrix, then extracts horizontal segments from the edge parameters. Since the polygons in the mask are Manhattan rectangles and closed shapes, the algorithm only needs to process segments along one direction (either horizontal or vertical), reducing the computational cost by half. For each segment, the algorithm performs parallel computation across all grid points within the bounding box, calling the \texttt{check\_cross} function to determine ray-segment intersections. The \texttt{check\_cross} function uses cross products to efficiently check if a ray from a point intersects a segment. After processing all segments, the even-odd rule is applied to finalize the binary mask based on the parity of intersections at each point. The algorithm leverages parallel computation, efficient ray-segment intersection checks, and the properties of Manhattan rectangles to enable fast and accurate mask generation, making it suitable for use in the DiffOPC framework.

The \textbf{forward} pass of the rasterization process converts the edge parameters, represented as a tensor of shape $[N_s, 2, 2]$, into a mask tensor of shape $[W, H]$, where $N_s$ is the number of segments, and $W$ and $H$ are the width and height of the mask, respectively.
In contrast, the \textbf{backward} pass requires transforming the gradients from the lithography model, which are of shape $[W, H]$, into gradients for the segments, represented as a tensor of shape $[N_s, 2, 2]$.
To accomplish this, the algorithm first computes the gradient of the mask tensor with respect to the edge parameters using automatic differentiation. Let $\frac{\partial L}{\partial \boldsymbol{M}}$ be the gradient of the loss function $L$ with respect to the mask tensor $\boldsymbol{M}$, obtained from the lithography model.
The goal is to calculate $\frac{\partial L}{\partial \hat{\boldsymbol{S}}}$, the gradient of the loss function with respect to the edge parameters $\hat{\boldsymbol{S}}$.
Applying the chain rule, we have:
$$
  \frac{\partial L}{\partial \hat{\boldsymbol{S}}} = \frac{\partial L}{\partial \boldsymbol{M}} \cdot \frac{\partial \boldsymbol{M}}{\partial \hat{\boldsymbol{S}}}.
$$
The term $\frac{\partial \boldsymbol{M}}{\partial \hat{\boldsymbol{S}}}$ represents the Jacobian matrix of the rasterization process, which maps changes in edge parameters to changes in the mask tensor.
This Jacobian matrix is computed efficiently using \Cref{alg:edge_grad}.
In our implementation, as in the \texttt{Interpolate} function in \cref{line:interpolate}, we choose the gradient at the midpoint of each segment as the representative gradient for that segment, as stated in \cref{eq:gradient_interpolation}.
Once the Jacobian matrix is obtained, the gradient of the loss function with respect to the edge parameters can be calculated by multiplying the gradient of the loss function with respect to the mask tensor, $\frac{\partial L}{\partial \boldsymbol{M}}$, by the Jacobian matrix $\frac{\partial \boldsymbol{M}}{\partial \boldsymbol{S}}$. This operation effectively backpropagates the gradients from the lithography model to the edge parameters, enabling the optimization of the edge-based OPC problem using gradient-based methods.

\begin{algorithm}
\caption{Transform Mask to Edge Gradients with Velocity}
\begin{algorithmic}[1]
\State \textbf{Input:} Gradient matrix $\frac{\partial L}{\partial \boldsymbol{M}}$ of size $W \times H$;
\State \textbf{Input:} Edge segments $\hat{\boldsymbol{S}}$ of shape $[N_s, 2, 2]$, where each edge is defined by two points: start $(x_1, y_1)$ and end $(x_2, y_2)$;
\State \textbf{Input:} Pre-defined velocity list $\boldsymbol{V}$ for each segment $\boldsymbol{s}_i$;
\State \textbf{Output:} Edge gradients $\frac{\partial L}{\partial \hat{\boldsymbol{S}}}$ of shape $[N_s, 2, 2]$;
\Function{ComputeEdgeGradients}{$\frac{\partial L}{\partial \boldsymbol{M}}, \hat{\boldsymbol{S}}, \boldsymbol{V}$}
\State Initialize $\frac{\partial L}{\partial \hat{\boldsymbol{S}}}$ to zeros of shape $[N_s, 2, 2]$;
\For{each segment $i$ in $\hat{\boldsymbol{S}}$}
    \State $(v_x, v_y) \gets \boldsymbol{V}[i]$;
    \State $(m_x, m_y) \gets \left[({x_1 + x_2})/{2}, ({y_1 + y_2}/{2})\right]$; \Comment{Midpoint}
    \State $g_{\text{mid}} \gets \texttt{Interpolate}(\frac{\partial L}{\partial \boldsymbol{M}}, m_x, m_y)$; \label{line:interpolate}
    \State $\boldsymbol{v}_i \gets [[v_x, v_y],[v_x, v_y]]$; \Comment{Edge velocity}
    \State $\frac{\partial L}{\partial \hat{\boldsymbol{S}}}[i] \gets g_{\text{mid}} \cdot \boldsymbol{v}_i$;
\EndFor
\State \textbf{return} $\frac{\partial L}{\partial \hat{\boldsymbol{S}}}$;
\EndFunction
\end{algorithmic}
\label{alg:edge_grad}
\end{algorithm}

\minisection{MRC aware optimization} One of the significant advantages of EBOPC is the ability to obtain boundary information in real-time during the optimization process, including edges, line ends, jogs, notches, and other features.
This is not possible with PBOPC.
While level set-based methods can control boundaries globally, they lack the ability to fine-tune specific locations. 
DiffOPC generates MRC-clean optimization results by explicitly controlling manufacturability through the velocity term $\boldsymbol{v}_i$ during optimization.
Before the experiment, we divide the MRC edges into corresponding check pairs.
We classify mask rules into two categories: spacing checks, such as minimum spacing, end of line spacing, jog to jog spacing, and special notch spacing, and width checks, such as minimum width check.
Let $\boldsymbol{\delta}$ denote the distance vector between check pairs.
The projection of $\boldsymbol{\delta}$ along the y-direction is given by $\operatorname{proj}_y \boldsymbol{\delta} = (\boldsymbol{\delta} \cdot \boldsymbol{j}) \boldsymbol{j}$, where $\boldsymbol{j}$ is the unit vector in the y-direction. The projection along the x-direction is similarly defined. We achieve MRC-aware optimization by controlling the velocity $\boldsymbol{v}_i$ as follows:
$\boldsymbol{v}_i^{\prime} = \boldsymbol{v}_i \cdot \tau(\boldsymbol{\delta})$
where $\tau(\boldsymbol{\delta})$ is a function related to $\boldsymbol{\delta}$, defined as:
$\tau(\boldsymbol{\delta}) =  \sigma(\beta(\operatorname{proj}\boldsymbol{\delta} - D)).$
Here, $D$ is a constant related to the mask rule, and $\operatorname{proj}$ is the projection operator in either $x$ or $y$ direction, $\beta$ is the steepness of sigmoid function $\sigma(\cdot)$.
For the spacing and width check, when the distance $\operatorname{proj}\boldsymbol{\delta}$ is smaller than $D$, the velocity term rapidly decays to 0, preventing further reduction in the distance.
When $\operatorname{proj} \boldsymbol{\delta}$ is greater than $D$, $\tau(\boldsymbol{\delta})$ returns to 1, allowing normal optimization to proceed without interference.
By controlling the velocity term based on the distance between check pairs and mask rule constants, DiffOPC effectively incorporates MRC constraints into the optimization process.

\minisection{Lithography simulations}
After obtaining the mask $\boldsymbol{M}$ through the rasterization process, we can utilize forward lithography model in \cref{eq:approx_intensity} to calculate the aerial intensity $\boldsymbol{I}$. To obtain a continuous-valued printed image $\boldsymbol{Z}$, we employ the sigmoid function $\sigma(\cdot)$ to scale \cref{eqn:resist1} into a continuous space:
$\boldsymbol{Z} = \sigma(\alpha (\boldsymbol{I} - I_{th}))$,
where $\alpha$ is the steepness of $\sigma(\cdot)$, and $I_{th}$ is the threshold intensity value.

\minisection{Objective function}
We employ a combination of three loss functions: $L_2$ loss, PVB loss, and EPE loss.
The $L_2$ loss and PVB loss are defined as:
\begin{equation}
\mathcal{L}_2 = \Vert \boldsymbol{Z}_{nom} - \boldsymbol{T} \Vert ^2,\ \ \mathcal{L}_{pvb} = \Vert \boldsymbol{Z}_{max} - \boldsymbol{Z}_{min} \Vert ^2.
\end{equation}
For the EPE loss, measured points are sampled along the boundary of the target patterns, which includes a set of samples on horizontal edges (HS) and a set of samples on vertical edges (VS).
To map the EPE loss to the continuous-value domain, we utilize the sigmoid function.
First, we calculate the distance between $\boldsymbol{Z}_{nom}$ and the target pattern $\boldsymbol{T}$ at the sampled points in VS and HS:
\begin{equation}
\begin{aligned}
  \boldsymbol{D}_{sum_{ij}} &= \begin{cases}
\sum_{k=j-th_{epe}}^{j+th_{epe}} \boldsymbol{D}_{ik}, & \text{if } (i, j) \in \text{HS}, \\
\sum_{k=i-th_{epe}}^{i+th_{epe}} \boldsymbol{D}_{kj}, & \text{if } (i, j) \in \text{VS},
\end{cases}
\end{aligned}
\end{equation}
where $\boldsymbol{D}_{ik}$ and $\boldsymbol{D}_{kj}$ represent the distances between the printed image and the target pattern at the corresponding locations,
and $th_{epe}$ is a threshold value that determines the neighborhood size for the distance calculation.
$\boldsymbol{D}$ is calculated by $\boldsymbol{D} = (\boldsymbol{Z}_{nom} - \boldsymbol{T})^2$.
Next, we apply the sigmoid function to the calculated distances to obtain the continuous-valued EPE loss:
\begin{equation}
\mathcal{L}_{epe} = \sum_{(i, j) \in HS \cup VS} \frac{1}{1+\exp(-\gamma \boldsymbol{D}_{sum_{ij}})},
\label{eq:epe_loss}
\end{equation}
where $\gamma$ is a scaling factor that controls the steepness of the sigmoid function.
The total loss function is then defined as a weighted sum of the three individual loss components:
\begin{equation}
\mathcal{L}_{total} = w_1 \mathcal{L}_2 + w_2 \mathcal{L}_{pvb} + w_3 \mathcal{L}_{epe},
\end{equation}
where $w_1$, $w_2$, and $w_3$ are the weights assigned to each loss component.
The use of the sigmoid function in the EPE loss allows for a smooth integration of the EPE into the continuous-value domain, enabling efficient gradient-based optimization.

For the backward pass, the gradients of the total loss function with respect to the segment $\boldsymbol{s}_i$ are calculated using the chain rule:
\begin{equation}
  \begin{aligned}
    \frac{\partial \mathcal{L}}{\partial \boldsymbol{s}_i} &= \frac{\partial \mathcal{L}}{\partial \boldsymbol{M}} \cdot \frac{\partial \boldsymbol{M}}{\partial \boldsymbol{s}_i} = \frac{\partial \mathcal{L}}{\partial \boldsymbol{M}}\left[\lfloor\frac{x_{i1} + x_{i2}}{2}\rfloor, \lfloor\frac{y_{i1} + y_{i2}}{2}\rfloor\right]\cdot \boldsymbol{v}_i,
  \end{aligned}
  \label{eq:gradient_interpolation}
\end{equation}
where $\lfloor\cdot\rfloor$ is floor operation and
\begin{equation}
  \frac{\partial L}{\partial \boldsymbol{M}} = w_1\frac{\partial \mathcal{L}_{{2}}}{\partial \boldsymbol{M}} + w2\frac{\partial \mathcal{L}_{{pvb}}}{\partial \boldsymbol{M}} + w_3\frac{\partial \mathcal{L}_{{epe}}}{\partial \boldsymbol{M}}.
\end{equation}
For the $L_2$ loss, the gradient is calculated as:
\begin{equation}
  \begin{aligned}
  \frac{\partial \mathcal{L}_{{2}}}{\partial \boldsymbol{M}} =& 2\cdot\left(\boldsymbol{Z}-\boldsymbol{T}\right) \odot \frac{\partial \boldsymbol{Z}}{\partial \boldsymbol{M}} \\
  =&2\alpha\cdot \left\{\boldsymbol{H}^{\prime} \otimes\left[\left(\boldsymbol{Z}-\boldsymbol{T}\right) \odot \boldsymbol{Z} \odot(1-\boldsymbol{Z}) \odot\left(\boldsymbol{M} \otimes \boldsymbol{H}^{*}\right)\right]\right.\\
  &\left.+(\boldsymbol{H}^{{\prime}})^{*} \otimes\left[\left(\boldsymbol{Z}-\boldsymbol{T}\right) \odot \boldsymbol{Z} \odot(1-\boldsymbol{Z}) \odot(\boldsymbol{M} \otimes \boldsymbol{H})\right]\right\},
  \end{aligned}
  \label{eq:l2_gradient}
\end{equation}
where the $\boldsymbol{H}^{\prime}$ is the flipped kernel set $\boldsymbol{H}$, and the $\boldsymbol{H}^{*}$ is the conjugate of $\boldsymbol{H}$.
Similarly, for the PVB loss, the gradient is calculated as:
\begin{equation}
  \frac{\partial L_{{pvb}}}{\partial {\boldsymbol{M}}}=2 \times\left(\boldsymbol{Z}_{{min}}-\boldsymbol{Z}_{max}\right) \odot\left(\frac{\partial {\boldsymbol{Z}_{{min}}}}{\partial \boldsymbol{M}}-\frac{\partial {\boldsymbol{Z}_{{max}}}}{\partial \boldsymbol{M}}\right) .
  \label{eq:pvb_gradient}
\end{equation}
The derivation of $\frac{\partial {\boldsymbol{Z}_{{min}}}}{\partial \boldsymbol{M}}$ and $\frac{\partial {\boldsymbol{Z}_{{max}}}}{\partial \boldsymbol{M}}$ is similar to that of $\frac{\partial \boldsymbol{Z}}{\partial \boldsymbol{M}}$ in \cref{eq:l2_gradient}.
For the EPE loss, the gradient is calculated by summarizing the gradients at the measure points $(i, j)$:
\begin{equation}
  \frac{\partial \mathcal{L}_{epe}}{\partial \boldsymbol{M}} = \sum_{(i, j) \in HS \cup VS} \frac{\partial \mathcal{L}_{epe}}{\partial \boldsymbol{D}_{sum_{ij}}} \cdot \frac{\partial \boldsymbol{D}_{sum_{ij}}}{\partial \boldsymbol{M}},
\end{equation}
where
\begin{equation}
  \resizebox{.91\linewidth}{!}{$
  \frac{\partial \mathcal{L}_{epe}}{\partial \boldsymbol{D}_{sum_{ij}}} = \gamma \cdot \frac{1}{1+\exp(-\gamma \boldsymbol{D}_{sum_{ij}})} (1 - \frac{1}{1+\exp(-\gamma \boldsymbol{D}_{sum_{ij}})}),
  $}
\end{equation}
and
\begin{equation}
  \frac{\partial  \boldsymbol{D}_{sum_{ij}}}{\partial \boldsymbol{M}} = \begin{cases}
    \sum_{k=j-th_{epe}}^{j+th_{epe}} \frac{\partial \boldsymbol{D}_{ik}}{\partial \boldsymbol{M}}, & \text{if } (i, j) \in \text{HS}, \\
    \sum_{k=i-th_{epe}}^{i+th_{epe}} \frac{\partial \boldsymbol{D}_{kj}}{\partial \boldsymbol{M}}, & \text{if } (i, j) \in \text{VS},
    \end{cases}
\end{equation}
with $\frac{\partial \boldsymbol{D}_{ij}}{\partial \boldsymbol{M}}$ calculated as:
\begin{equation}
  \begin{aligned}
  \frac{\partial D_{i j}}{\partial \boldsymbol{M}} &= \frac{\partial\left(\boldsymbol{Z}_{ij}-\boldsymbol{T}_{ij}\right)^2}{\partial \boldsymbol{M}}  = 2 (\boldsymbol{Z}_{ij} - \boldsymbol{T}_{ij}) \cdot \frac{\partial \boldsymbol{Z}_{ij}}{\partial \boldsymbol{M}}.\\
  \end{aligned}
\end{equation}
The detailed derivation of $\frac{\partial \boldsymbol{Z}_{ij}}{\partial \boldsymbol{M}}$ can be found in \cref{eq:l2_gradient}.

\minisection{SRAF generation}
SRAFs in lithography enhance sub-resolution element printability by modifying diffraction and interference patterns in photoresist, leading to widened process windows, improved resolution, depth of focus, and reduced line edge roughness.
The primary distinction among prior works lies in their handling of SRAFs. In level set-based ILT methods, the implicit function \(\phi\) is tied to the primary pattern, preventing the generation of SRAFs during optimization. Conversely, pixel-based ILT methods like~\cite{OPC-multiilt2023sun} can generate SRAFs during mask optimization due to their higher degree of freedom. However, pixel-based ILT cannot impose rule-based constraints on SRAFs, causing their growth to rely solely on gradients. This improves printability but can increase MRC violations and hotspots.
To address these issues, we propose a two-stage SRAF optimization algorithm. The first stage involves efficient SRAF seed generation using gradient contours, and the second stage employs a differentiable edge-based optimization for the generated SRAFs. This approach effectively avoids the problems of missing SRAFs in level set-based methods and violations in pixel-based SRAFs.

\begin{figure}
  \centering
  \subfloat[]{\includegraphics[width=.42\linewidth]{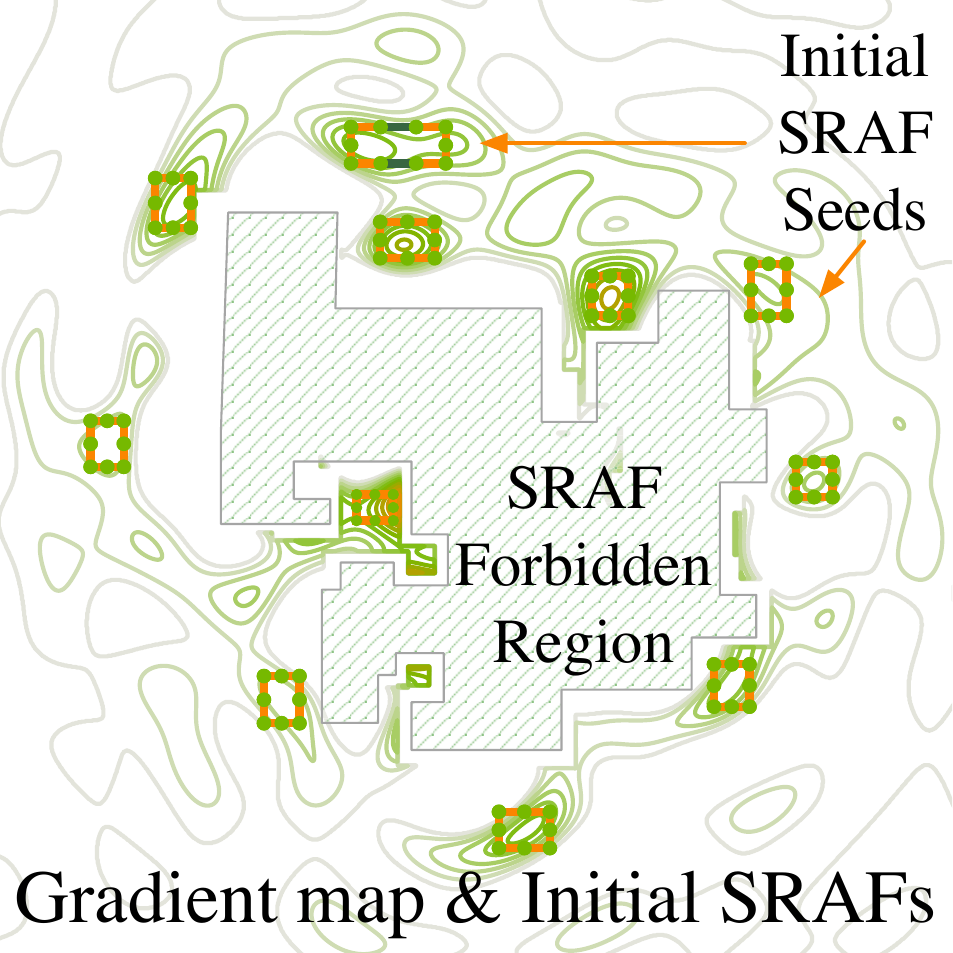}\label{fig:initialsraf}} \hspace{1em}
  \subfloat[]{\includegraphics[width=.42\linewidth]{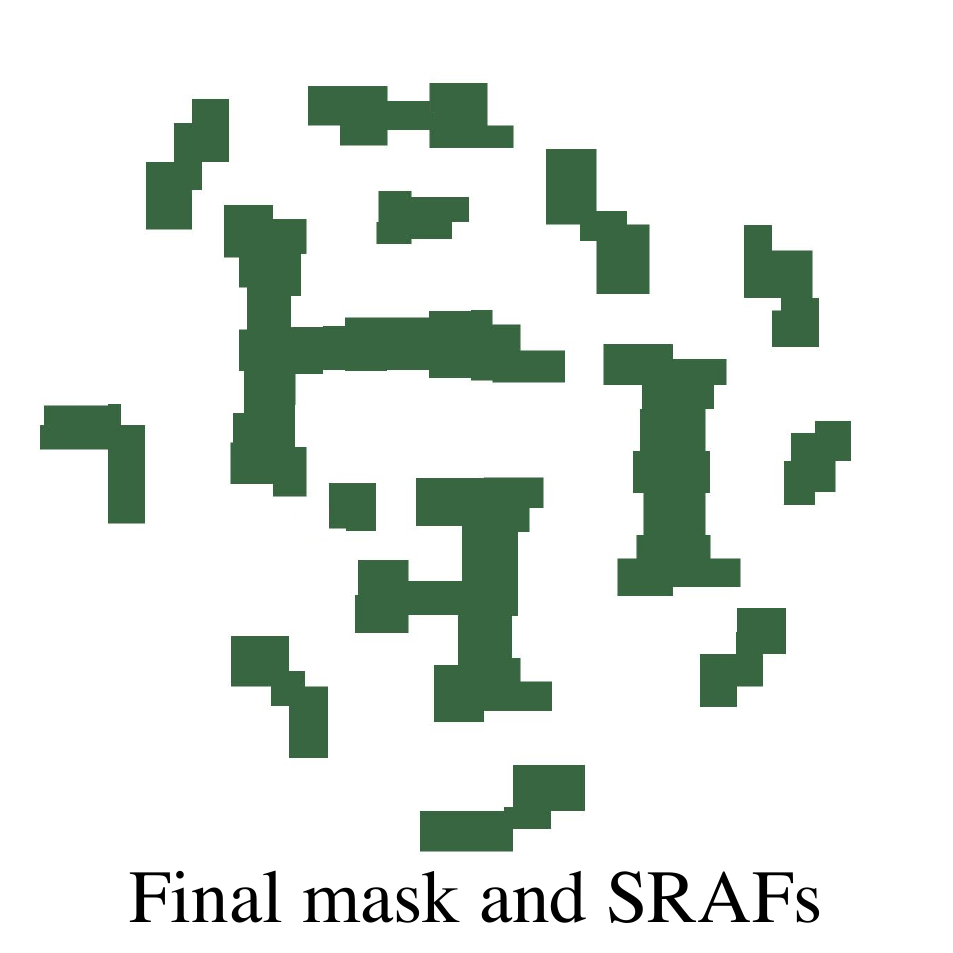}\label{fig:finalsraf}}
  \caption{DiffOPC SRAF insertion and optimization.}
  \label{fig:sraf}
\end{figure}

\textit{Gradient Contour-based SRAF Seed Generation:}
During the optimization process, we observe that certain regions near the main pattern exhibit gradients that flip the mask value, changing it from 0 to 1. However, since the edge-based segments do not include these regions, they remain at 0. Combining continuous transmission mask (CTM)~\cite{ctmsraf-yu2023} theory and the results from~\cite{OPC-multiilt2023sun}, we conclude that these gradients can contribute to SRAF generation.
As depicted in \Cref{fig:initialsraf}, the contour line of the mask gradient map shows the position of the extreme gradient points and indicates the gradient drop rate. The position of the extreme points can guide SRAF placement, while the gradient information can guide the subsequent SRAF cleanup process.
The implementation involves extending the existing mask by a certain distance related to the mask rules to create a SRAF forbidden region. As illustrated in \Cref{fig:initialsraf}, gradient contour lines are drawn outside the SRAF forbidden region. The extreme points and the corresponding contour aspect ratios are used as the center of the SRAF seeds. The initial SRAF minimum width/length is set to a fixed value, and the shape and placement of the SRAF are determined based on the aspect ratio. This step does not require precise SRAF generation; it only needs to determine the initial position and aspect ratio.

\textit{Differentiable Edge-based SRAF Optimization:}
In the second stage, the generated seeds illustrated in \Cref{fig:initialsraf} are processed using the \Cref{alg:edge_init} segmentation method to create new segments, which are then added to the main optimization process. The SRAFs are optimized together with the mask.
To accelerate the SRAF optimization process, we adopt a multi-resolution strategy similar to~\cite{OPC-multiilt2023sun}. SRAF seeds are generated at low resolution, and then the seeds and mask are refined in high resolution for more precise optimization. Sample results are shown in \Cref{fig:finalsraf}. The proposed two-stage SRAF optimization algorithm enables the generation of SRAFs that enhance printability while minimizing MRC violations.
We conducted a comprehensive comparison of DiffOPC, ILTs, and EBOPC in \Cref{tab:cmp_all_methods}.

\begin{table*}[!htp]\centering
  \caption{DiffOPC compared with ILTs and EBOPC.}\label{tab:cmp_all_methods}
  \begin{tabular}{lllll}\toprule
  &Level set-based ILT &Pixel-based ILT &MEEF-based EBOPC &DiffOPC \\\midrule
  Shape &Free-form (more smooth) &Free-form (sharp corners) &Manhattan shape &Manhattan shape \\
  Gradient update: &Level set function &Mask transmission &Edge move distance &Edge segment position \\
  Optimization &Gradient descent &Gradient descent &Newton's method &Gradient descent \\
  SRAF &None &Automatic generation &Pre-placed SRAF &SRAF co-optimization \\
  MRC clean? &No &No &Yes &Yes \\
  EPE loss &None &None &Yes &Yes \\
  \bottomrule
  \end{tabular}
\end{table*}

\section{Experimental Results}
\label{sec:exp}
In our implementation, we set $N_k = 24$ for the SOCS approximation. The parameters $\alpha = \beta = \gamma = 50$, $w_1=1, w_2=0.9, w_3=100$. The default segment length is set to 80 nm. The lithography recipe is provided by the ICCAD 2013~\cite{OPC-ICCAD2013-Banerjee} contest evaluation package. The mask fracturing tool is implemented based on a GPU-accelerated rectangular decomposition algorithm~\cite{gpu-rec-decomposition}. The entire framework is written in PyTorch and tested on an Nvidia RTX 3090 GPU. The mask rule check (MRC) is performed using KLayout.
DiffOPC is tested on both metal layer and via layer designs. The metal layer evaluation designs for 32 nm M1 layout designs are from~\cite{OPC-ICCAD2013-Banerjee}, and larger layouts from~\cite{OPC-multiilt2023sun} for the same process node. The via layer evaluation designs are adopted from~\cite{zheng2023lithobench}, containing ten $2 \mu m \times 2 \mu m$ clips with different numbers of $70 nm \times 70 nm$ via patterns. SRAF seeds are generated in a low resolution of $512 \times 512$ and optimized at a resolution of $2048 \times 2048$.
\begin{table*}[!tbp]\centering
  \caption{Comparison with ILT methods on ICCAD13 dataset.}\label{tab:res_iccad13}
  \setlength{\tabcolsep}{3pt}
  \resizebox{.99\linewidth}{!}{
  \begin{tabular}{cc|ccccc|ccccc|ccccc|ccccc}\toprule
  & &\multicolumn{5}{c|}{\texttt{ICCAD'20} NeuralILT~\cite{OPC-ICCAD2020-NeuralILT}} &\multicolumn{5}{c|}{\texttt{DAC'23} MultiILT~\cite{OPC-multiilt2023sun}} &\multicolumn{5}{c|}{MultiILT (Post-MRC)~\cite{OPC-multiilt2023sun}} &\multicolumn{5}{c}{DiffOPC} \\
  & &L2 &PVB &EPE &\#shots &TAT &L2 &PVB &EPE &\#shots &TAT &L2 &PVB &EPE &\#shots &TAT &L2 &PVB &EPE &\#shots &TAT \\\midrule
  c1 &215344 &50795 &63695 &8 &743 &13.57 &40779 &50661 &3 &307 &3.49 &45940 &54949 &7 &275 &18.42 &38661 &55156 &3 &107 &10.62 \\
c2 &169280 &36969 &60232 &3 &571 &14.37 &34201 &44322 &2 &186 &3.47 &37035 &45085 &3 &167 &13.64 &29548 &45610 &0 &104 &10.65 \\
c3 &213504 &94447 &85358 &52 &791 &9.72 &66486 &71527 &22 &308 &3.47 &79751 &82213 &35 &261 &18.71 &64706 &93773 &19 &121 &11.52 \\
c4 &82560 &17420 &32287 &2 &209 &10.40 &10942 &21500 &0 &233 &3.47 &13111 &32330 &1 &204 &19.24 &12054 &25053 &0 &80 &6.04 \\
c5 &281958 &42337 &65536 &3 &631 &10.04 &30231 &51277 &0 &374 &3.47 &39236 &60069 &1 &296 &13.23 &31774 &56966 &0 &129 &6.72 \\
c6 &286234 &39601 &59247 &5 &745 &11.11 &30741 &44982 &0 &365 &3.47 &37493 &56581 &1 &300 &16.14 &31791 &52997 &0 &129 &10.33 \\
c7 &229149 &25424 &50109 &0 &354 &9.67 &17101 &40294 &0 &196 &3.50 &19133 &48156 &0 &155 &13.57 &17847 &45791 &0 &96 &6.59 \\
c8 &128544 &15588 &25826 &0 &467 &11.81 &11935 &20357 &0 &243 &3.47 &13917 &28910 &0 &201 &22.09 &11641 &23172 &0 &78 &6.52 \\
c9 &317581 &52304 &68650 &2 &653 &9.68 &35805 &57930 &0 &435 &3.50 &45659 &70023 &1 &387 &14.51 &36595 &65732 &0 &141 &10.11 \\
c10 &102400 &10153 &22443 &0 &423 &11.46 &8825 &18470 &0 &114 &3.48 &9715 &22979 &0 &88 &18.23 &8184 &17923 &0 &76 &5.12 \\\midrule
\multicolumn{2}{c|}{Average} &38503.8 &53338.3 &7.5 &558.7 &11.18 &28704.6 &42132.0 &2.7 &276.1 &3.48 &34099 &50130 &4.9 &233.4 &16.78 &\textbf{28280} &48217 &\textbf{2.2} &\textbf{106.1} &8.42 \\
\multicolumn{2}{c|}{Ratio} &1.36 &1.11 &3.41 &5.27 &1.33 &1.02 &0.87 &1.23 &2.60 &0.41 &1.21 &1.04 &2.23 &2.20 &1.99 &\textbf{1.00} &1.00 &\textbf{1.00} &\textbf{1.00} &1.00 \\
  \bottomrule
  \end{tabular}}
\end{table*}
\begin{table*}[!htp]\centering
  \vspace{-.5em}
  \caption{Comparison with ILT methods on larger dataset.}\label{tab:res_large}
  \setlength{\tabcolsep}{3pt}
  \resizebox{.99\linewidth}{!}{
  \begin{tabular}{cc|ccccc|ccccc|ccccc|ccccc}\toprule
  & &\multicolumn{5}{c|}{\texttt{ICCAD'20} NeuralILT~\cite{OPC-ICCAD2020-NeuralILT}} &\multicolumn{5}{c|}{\texttt{DAC'23} MultiILT~\cite{OPC-multiilt2023sun}} &\multicolumn{5}{c|}{MultiILT (Post-MRC)~\cite{OPC-multiilt2023sun}} &\multicolumn{5}{c}{DiffOPC} \\
  & &L2 &PVB &EPE &\#shots &TAT &L2 &PVB &EPE &\#shots &TAT &L2 &PVB &EPE &\#shots &TAT &L2 &PVB &EPE &\#shots &TAT \\\midrule
  L1 &494560 &79933 &120577 &12 &669 &20 &64020 &93060 &3 &628 &3.48 &80403 &101194 &11 &529 &22.66 &57178 &97979 &3 &247 &11.49 \\
  L2 &448496 &86995 &104266 &15 &556 &12 &52072 &84733 &1 &553 &3.46 &72261 &91673 &8 &491 &18.07 &63288 &85388 &2 &109 &11.89 \\
  L3 &492720 &133281 &152718 &70 &766 &15 &95174 &116687 &30 &641 &3.49 &118860 &125013 &65 &564 &20.49 &81120 &120828 &22 &267 &14.69 \\
  L4 &361776 &43797 &92137 &0 &455 &14 &33076 &67839 &1 &523 &3.47 &41526 &76582 &2 &479 &21.43 &31531 &70713 &0 &177 &8.53 \\
  L5 &561174 &69521 &122115 &3 &808 &19 &55013 &100120 &0 &670 &3.46 &76176 &111861 &6 &528 &16.00 &53484 &102675 &0 &258 &7.69 \\
  L6 &565450 &73790 &117359 &2 &764 &19 &57386 &94863 &0 &670 &3.45 &76644 &108667 &5 &501 &18.74 &56581 &97980 &0 &293 &13.21 \\
  L7 &445365 &49031 &92320 &0 &531 &19 &32947 &73799 &0 &648 &3.45 &40838 &84006 &0 &503 &18.19 &42091 &84836 &0 &222 &9.91 \\
  L8 &407760 &47409 &84971 &0 &478 &16 &41265 &67797 &0 &493 &3.48 &43475 &73021 &0 &426 &20.49 &32482 &68687 &0 &198 &8.53 \\
  L9 &596797 &93922 &115028 &5 &614 &14 &70385 &108998 &0 &541 &3.48 &84857 &120426 &4 &514 &18.04 &60748 &111449 &0 &226 &11.44 \\
  L10 &381616 &28028 &80127 &0 &452 &19 &30091 &62206 &0 &546 &3.46 &36767 &67807 &0 &452 &20.95 &28334 &63274 &0 &188 &8.78 \\\midrule
  \multicolumn{2}{c|}{Average} &70570.7 &108161.8 &10.7 &609.3 &16.7 &53142.9 &87010.2 &3.5 &591.3 &3.47 &67181 &96025 &10.1 &498.7 &19.51 &\textbf{50684} &90381 &\textbf{2.7} &\textbf{218.5} &10.62 \\
  \multicolumn{2}{c|}{Ratio} &1.39 &1.20 &3.96 &2.79 &1.57 &1.05 &0.96 &1.30 &2.71 &0.33 &1.33 &1.06 &3.74 &2.28 &1.84 &\textbf{1.00} &1.00 &\textbf{1.00} &\textbf{1.00} &1.00 \\
  \bottomrule
  \end{tabular}}
\end{table*}
\subsection{Experimental Results on Metal Layer}
\begin{figure}
  \centering
\definecolor{ICCAD13Color}{HTML}{76b900}
\definecolor{LargeColor}{HTML}{fb8500}
\definecolor{ViaColor}{HTML}{ffe0a2}

\begin{tikzpicture}
    \begin{axis}[
        ybar,
        bar width=.5cm, 
        width=\linewidth,
        height=.4\linewidth,
        enlarge x limits=0.3, 
        symbolic x coords={A, B, C},
        xticklabels={%
          NeuralILT~\cite{OPC-ICCAD2020-NeuralILT},
          MultiILT~\cite{OPC-multiilt2023sun},
          DiffOPC 
        },
        xtick=data,
        ylabel={Avg. MRC violations},
        ylabel style={font=\scriptsize},
        nodes near coords,
        ymin=0,
        ymax=35, 
        legend image code/.code={
          \draw [#1] (0cm,-0.1cm) rectangle (0.1cm,0.2cm);
        },
        legend cell align={left},
        legend style={
          at={(0.82,0.88)},
          draw=none,
          fill=none,
          anchor=north,
          legend columns=1},
    ]
    \addplot[fill=ICCAD13Color] coordinates {(A, 8.9)  (B, 11.7) (C, 0)};
    \addplot[fill=LargeColor] coordinates {(A, 18.6) (B, 25.4) (C, 0)};
    \addplot[fill=ViaColor] coordinates {(A, 4)    (B, 22) (C, 0)};
    \legend{ICCAD13, Large Dataset, Via}
    \end{axis}
\end{tikzpicture}
  \caption{MRC violations across methods and datasets}
  \label{fig:mrc}
  \vspace{-0.2em}
\end{figure}
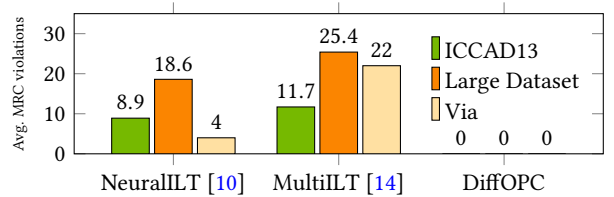

\minisection{Comparison with ILT}
\Cref{tab:res_iccad13} compares the performance of our proposed DiffOPC framework with state-of-the-art (SOTA) ILT approaches, namely NeuralILT~\cite{OPC-ICCAD2020-NeuralILT} and MultiILT~\cite{OPC-multiilt2023sun}, on the ICCAD13 benchmark.
The comparison is based on key metrics such as L2 ($nm^2$), PVB ($nm^2$), EPE ($nm$), number of shots, and turnaround time (TAT, seconds).
DiffOPC demonstrates superior performance, achieving an average L2 of 28280, which is $1.5\%$ and $27\%$ lower than MultiILT and NeuralILT, respectively.
Attributed to the utilization of EPE loss introduced in \cref{eq:epe_loss}, DiffOPC achieves lower EPE, with an average of 2.2, representing a $19\%$ and $71\%$ reduction compared to MultiILT and NeuralILT.
Moreover, DiffOPC requires significantly fewer shots per case, with an average of 106.1 shots, representing a $62\%$ and $81\%$ reduction compared to MultiILT and NeuralILT, which translates to lower manufacturing costs.
These results highlight the effectiveness of DiffOPC in generating mask patterns with improved printability while maintaining better manufacturability compared to ILT methods.
As mentioned in \Cref{sec:intro} and illustrated in \Cref{fig:mrc}, ILT approaches are prone to introducing MRC violations, which do not meet industrial requirements.
We also present the post-MRC results for MultiILT in the "Post-MRC" column, where the TAT includes both the ILT runtime and the post-processing time for cleaning mask rule violations.
It is noteworthy that the post-MRC stage for MultiILT leads to a significant performance degradation, evident from the increased average values of L2, PVB, EPE, and TAT compared to the original MultiILT results.
The results of DiffOPC outperform all metrics of ILT in the post-MRC stage.
This indicates that ILT-generated patterns may not optimize as desired and could introduce more violations, prolonging processing times due to MRC.
In contrast, DiffOPC maintains superior performance without extra post-processing steps, highlighting its robustness and efficiency in generating high-quality, manufacturable mask patterns meeting industrial standards.

\minisection{Large dataset}
To further validate the robustness and scalability of our proposed DiffOPC framework, we conduct experiments on a larger dataset and compare its performance with SOTA methods in \Cref{tab:res_large}.
The results demonstrate that DiffOPC consistently outperforms the other methods, highlighting its effectiveness in handling complex and diverse patterns.
DiffOPC achieves an average L2 of 50684, which is $4.7\%$ and $28.2\%$ lower than MultiILT and NeuralILT, respectively. Moreover, it exhibits superior performance in terms of EPE, with an average EPE of 2.7,
representing a $23\%$ and $75\%$ reduction over MultiILT and NeuralILT.
Notably, DiffOPC requires significantly fewer shots per case, with an average of 218.5 shots, which is $63\%$ and $64\%$ lower than MultiILT and NeuralILT.
As observed in the previous experiment, the post-MRC stage for MultiILT leads to a deterioration in performance.
This further underscores the limitations of ILT-based methods in generating manufacturable patterns that comply with industrial requirements.

\minisection{Comparison with MEEF-based EBOPC on ICCAD2013 benchmark}
To provide a fair comparison between proposed DiffOPC and the traditional MEEF-based EBOPC method~\cite{MEEF-lei2014model},
we evaluate both approaches on GPU without the inclusion of SRAFs.
The results in \Cref{tab:meef_res} demonstrate that DiffOPC consistently outperforms MEEF-EBOPC.
On average, DiffOPC achieves an L2 of 30579.6, which is $6.5\%$ lower than MEEF-EBOPC.
Similarly, DiffOPC exhibits lower average PVB, EPE, and TAT with $3.9\%$, $58\%$, and $59\%$ respectively.
These findings highlight the superior printability of the mask patterns generated by DiffOPC compared to the traditional MEEF-EBOPC.
It is worth noting that MEEF-EBOPC struggles to handle complex patterns.
The limitation is evident from the results presented in \Cref{tab:meef_res}, where MEEF-EBOPC exhibits particularly high EPE values for complex test cases such as c3 (EPE = 59) compared to simpler cases like c10 (EPE = 0).
In contrast, DiffOPC demonstrates robust performance across all test cases while maintaining a competitive shot count compared to MEEF-EBOPC.
\begin{table}[!tp]\centering
  \caption{Comparison with traditional MEEF EBOPC on ICCAD 2013 benchmark.}\label{tab:meef_res}
  \setlength{\tabcolsep}{2pt}
  \resizebox{.99\linewidth}{!}{
  \begin{tabular}{c|ccccc|ccccc}\toprule
  &\multicolumn{5}{c|}{MEEF-based EBOPC~\cite{MEEF-lei2014model}} &\multicolumn{5}{c}{DiffOPC w./o. SRAFs} \\
  &L2 &PVB &EPE &\#shots &TAT &L2 &PVB &EPE &\#shots &TAT \\\midrule
  c1 &52310 &60296 &14 &67 &13 &42177 &57981 &4 &79 &5.53 \\
  c2 &36498 &52124 &2 &60 &11 &31198 &50474 &2 &58 &5.35 \\
  c3 &90824 &103100 &59 &87 &12 &71643 &81219 &26 &92 &6.52 \\
  c4 &12144 &30663 &2 &34 &9 &14771 &32059 &0 &30 &3.29 \\
  c5 &31832 &60792 &0 &84 &14 &33986 &61796 &0 &89 &5.24 \\
  c6 &30612 &55751 &0 &98 &14 &33578 &56752 &0 &85 &5.44 \\
  c7 &15343 &48968 &0 &59 &11 &17928 &48886 &0 &60 &4.16 \\
  c8 &11851 &26149 &0 &33 &9 &12805 &25942 &0 &43 &3.82 \\
  c9 &38858 &71288 &0 &93 &14 &39543 &73183 &0 &97 &4.70 \\
  c10 &6562 &21024 &0 &26 &9 &8167 &21332 &0 &19 &3.39 \\
  Avg. &32683.4 &53015.5 &7.7 &64.1 &11.6 &\textbf{30579.6} &\textbf{50962.4} &\textbf{3.2} &{65.2} &\textbf{4.74} \\
  Ratio &1.07 &1.04 &2.33 &0.98 &2.44 &\textbf{1.00} &\textbf{1.00} &\textbf{1.00} &{1.00} &\textbf{1.00} \\
  \bottomrule
  \end{tabular}}
\end{table}

\subsection{Experimental Results on Via Layer}
In \Cref{tab:via_res}, we evaluate the performance of DiffOPC on the via layer against SOTA ILT and EBOPC methods, including a commercial tool, Calibre~\cite{TOOL-calibre}.

\minisection{Comparison with ILT methods}
DiffOPC outperforms ILT methods in terms of L2 and EPE, achieving the lowest values of 3957 and 13.5, respectively.
Notably, DiffOPC achieves these improvements while maintaining a significantly lower shot count (9.7 shots on average), which is approximately 1/20th of the shot count required by ~\cite{OPC-multiilt2023sun} (225 shots).

\minisection{Comparison with EBOPC methods}
Among the EBOPC methods, DiffOPC demonstrates superior performance, achieving the lowest L2 (3957), EPE (13.5), and TAT (2.8 seconds) compared to the commercial Calibre tool and the MEEF-based approach.
\begin{table}[!tbp]\centering
  \caption{Result comparison on via layer.}
  \label{tab:via_res}
  \setlength{\tabcolsep}{3pt}
  \resizebox{\linewidth}{!}{
  \begin{tabular}{c|ccc|ccc}\toprule
  &\multicolumn{3}{c|}{ILT} &\multicolumn{3}{c}{EBOPC} \\
  &NILT~\cite{OPC-ICCAD2020-NeuralILT} &MILT~\cite{OPC-multiilt2023sun} &MILT(MRC)~\cite{OPC-multiilt2023sun} &Calibre~\cite{TOOL-calibre} &MEEF~\cite{MEEF-lei2014model} &DiffOPC \\\midrule
  L2 &4629 &3963 &4385 &4136 &4371 &\textbf{3957} \\
  PVB &11367 &10478 &11157 &10648 &11272 &10880 \\
  $^*$EPE &22 &14.2 &18.7 &14.2 &18.2 &\textbf{13.5} \\
  Shots &219 &225 &191 &8.5 &6.2  &9.7 \\
  TAT &5.7 &1.5 &5.4 &8.2 &4.6    &2.8 \\
  \bottomrule
  \multicolumn{4}{l}{\textit{\footnotesize{$^{*}$EPE: EPE threshold set to 1 $nm$}.}}
  \end{tabular}}
\end{table}

\subsection{Ablation Study}
\minisection{Efficiency of CUDA-accelerated ray casting rasterization}
We compare the runtime of our CUDA-accelerated ray casting rasterization approach with the traditional EBOPC method based on indexing and the find-then-move strategy. For a clip size of $2\mu m \times 2\mu m$, a single forward rasterization step in DiffOPC takes 16 milliseconds, while the traditional method requires 196 milliseconds, representing a $\bm{12.3}\times$ speedup achieved by our CUDA ray casting implementation.

\minisection{Ablation Study on Segment Length}
Segment length in DiffOPC also impacts optimization performance.
In an ablation study using the ICCAD 2013 benchmark, segment lengths of 60$nm$, 80$nm$, and 100$nm$ resulted in EPE of 2.6, 2.2, and 2.8, with runtimes of 8.95, 8.42, and 6.92 seconds.
This shows that optimal segment length selection can enhance OPC performance.
Future work could explore adaptive segment length strategies, adjusting lengths based on pattern complexity and optimization progress for better performance.

\subsection{Summary of Experimental Results}

The experimental results on both metal and via layers demonstrate DiffOPC's superiority over SOTA ILT, post-MRC ILT and EBOPC methods in terms of printability, manufacturability, and cost-efficiency.
On metal layers, DiffOPC consistently outperforms SOTA ILT methods, exhibiting reduced EPE and shot count, along with lower manufacturing costs, while maintaining competitive TAT. The proposed framework eliminates the need for additional post-processing to address MRC violations, making it an efficient and reliable edge-based OPC solution for large-scale OPC tasks.
On via layers, DiffOPC achieves the best performance in L2, EPE, and TAT among EBOPC methods, surpassing even the commercial Calibre tool. Compared to ILT methods, DiffOPC shows the lowest L2 and EPE values while significantly reducing the number of shots, leading to lower manufacturing costs and improved throughput.
These results highlight DiffOPC's enhanced printability, pattern fidelity, and computational efficiency.

\section{Conclusion}
We propose DiffOPC, a differentiable edge-based OPC framework that bridges the gap between the superior manufacturability of EBOPC and the enhanced performance of ILT.
By leveraging a CUDA-accelerated ray casting algorithm, DiffOPC enables a differentiable rasterization process that allows gradients to propagate through the lithography model, facilitating the efficient optimization of edge segment positions. This innovative approach results in significant improvements in key metrics such as L2 and EPE while maintaining an exceptionally low shot count, leading to substantially reduced manufacturing costs.
Moreover, DiffOPC incorporates an efficient SRAF generation method, which seamlessly integrates SRAF with the main pattern optimization for a holistic and effective OPC solution.
Experimental results highlight DiffOPC's superior performance and efficiency over SOTA EBOPC and ILT methods, making it a promising advancement in semiconductor technologies.
\clearpage
{
\bibliographystyle{IEEEtran}
\bibliography{ref/Top-sim,ref/DL,ref/DFM,ref/NeRF,ref/bench,ref/CV,ref/SMO,ref/bilevel,ref/DiffOPC}
}
\end{document}